\documentclass[10pt,twocolumn,letterpaper]{article}

\usepackage[pagenumbers]{cvpr} 
\usepackage{pifont}
\usepackage{balance}
\usepackage{multicol}
\usepackage{multirow}
\usepackage{graphicx}
\usepackage{makecell} 
\usepackage{caption} 
\usepackage{xcolor}
\usepackage{colortbl}
\definecolor{mygray}{gray}{.9}

\usepackage[utf8]{inputenc}
\usepackage{booktabs}   
\usepackage{tabularx}   
\usepackage{xcolor}     
\usepackage{colortbl}   
\definecolor{lightgray}{gray}{0.9}
\usepackage{lineno}

\definecolor{cvprblue}{rgb}{0.21,0.49,0.74}
\usepackage[pagebackref,breaklinks,colorlinks,allcolors=cvprblue]{hyperref}

\def\paperID{****} 
\def\confName{CVPR}
\def\confYear{2026}

\title{
{TimeViper: A Hybrid Mamba-Transformer Vision-Language Model for \\ Efficient Long Video Understanding}
}

\author{Boshen Xu$^{1*\ddagger}$\quad Zihan Xiao$^{1*}$\quad Jiaze Li$^{2}$\quad Jianzhong Ju$^{2}$\quad Zhenbo Luo$^{2}$\quad Jian Luan$^{2}$\quad Qin Jin$^{1\dagger}$\\
$^1$ AIM3 Lab, Renmin University of China\quad 
$^2$ MiLM Plus, Xiaomi Inc.\\
{\small Project Page: \url{https://xuboshen.github.io/TimeViper/}}
}

\begin{document}

\twocolumn[{%
\renewcommand\twocolumn[1][]{#1}%
\maketitle
\begin{center}
\newcommand{\teaserwidth}{\textwidth}
\vspace{-0.35in}
    \centerline{        
        \includegraphics[width=\linewidth]{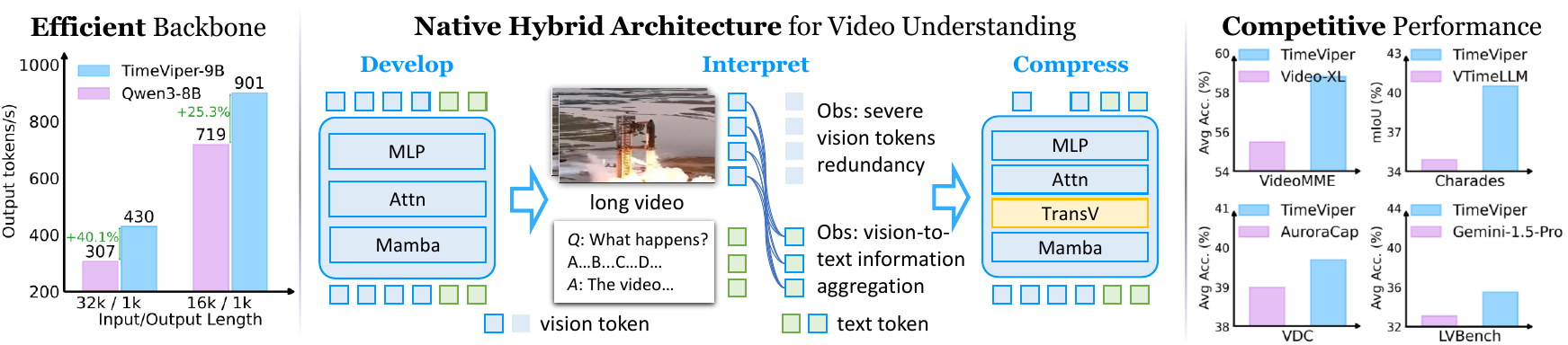}
    }
  \captionof{figure}{
We present \textbf{TimeViper}, a hybrid Mamba-Transformer vision-language model for efficient long video understanding. 
We reveal the severe vision token redundancy and a vision-to-text information aggregation phenomenon in hybrid models.
To this end, we introduce \textbf{TransV}, the first token-transfer module that compresses vision tokens into text tokens inside the LLM, enabling the model to process over 10,000 frames.
Benefitting from the Mamba layers' $O(n)$ computation and $O(1)$ cache cost, TimeViper generates 40.1\% more tokens per second than Qwen3~\cite{yang2025qwen3} when processing 32k input tokens (approximately 2k frames at 16 tokens per frame) and producing 1k output tokens with batch size 32. 
TimeViper delivers performance competitive with Transformer-based MLLMs on public benchmarks, including multi-choice QA on VideoMME~\cite{fu2025videomme} (vs. Video-XL~\cite{shu2025videoxl1}), temporal video grounding on Charades~\cite{sigurdsson2018charades} (vs. VTimeLLM~\cite{huang2024vtimellm}), video detailed captioning on VDC~\cite{chai2025auroracap} (vs. AuroraCap~\cite{chai2025auroracap}), and hour-long video understanding on LVBench~\cite{wang2025lvbench} (vs. Gemini-1.5-Pro~\cite{team2024gemini1.5pro}).
}
\label{fig:teaser}
\end{center}%
}]
\renewcommand{\thefootnote}%
{\fnsymbol{footnote}}
\footnotetext[0]{$\dagger$ Corresponding author; $\ddagger$ Project lead; $^*$ Equal contribution.\\ \textcolor{white}{spac}$^*$This work was done during their internship at Xiaomi.} 

\begin{abstract}
We introduce \textbf{TimeViper}, a hybrid vision-language model designed to tackle challenges of long video understanding.
Processing long videos demands both an efficient model architecture and an effective mechanism for handling extended temporal contexts.
To this end, TimeViper adopts a hybrid Mamba-Transformer backbone that combines the efficiency of state-space models with the expressivity of attention mechanisms.
Through this hybrid design, we reveal the \textit{vision-to-text information aggregation phenomenon}, where information progressively flows from vision tokens to text tokens across increasing LLM depth, resulting in \textit{severe vision token redundancy}. 
Motivated by this observation, we propose \textbf{TransV}, a token information transfer module that transfers and compresses vision tokens into instruction tokens while maintaining multimodal understanding capabilities.
This design enables TimeViper to process hour-long videos exceeding 10,000 frames.
Extensive experiments across multiple benchmarks demonstrate that TimeViper competes with state-of-the-art models while extending frame numbers. 
We further analyze attention behaviors of both Mamba and Transformer layers, offering new insights into hybrid model interpretability.  
This work represents an initial step towards developing, interpreting, and compressing hybrid Mamba-Transformer architectures.
\end{abstract}

\section{Introduction}
\label{sec:intro}

Understanding long videos is an essential yet long-standing challenge in computer vision, holding great potential for applications across video platforms~\cite{yang2023vidchapters,tapaswi2016movieqa}, household scenarios~\cite{perrett2025hdepic,yang2025egolife,grauman2022ego4d}, and embodied agents~\cite{zhang2024uninavid,black2024pi_0}.
Recent advances in multimodal large language models (MLLMs)~\cite{comanici2025gemini2.5pro} have made general long video understanding increasingly feasible.
Nevertheless, existing models still struggle to achieve a balance between effectiveness and efficiency when dealing with extended video contexts.
We argue that building a truly capable long-video understanding model requires addressing two key challenges:
\textit{constructing an efficient MLLM backbone}, and
\textit{handling redundancy in long-context processing}.

Most prior works~\cite{li2024videochatflash,shu2025videoxl1} adopt Transformer-based LLMs as the backbone, owing to their strong reasoning and language understanding capabilities.
However, the quadratic computational complexity of attention makes them inherently inefficient for long-context modeling.
To improve efficiency, recent efforts have explored linearized architectures such as Mamba~\cite{gu2024mamba,dao2024mamba2}, which replace attention with state-space models for linear-time inference. 
Despite their efficiency advantages, these models often depend heavily on distillation from Transformer-based models~\cite{li2025matvlm,liao2025mmmamba} or suffer from limited performance on complex multimodal tasks~\cite{wang2024longllava,zhao2025cobra}.
Encouragingly, a new generation of hybrid Mamba-Transformer LLMs~\cite{li2025minimax,basant2025nanov2,zuo2025falconh1,dong2025hymba,ren2025samba,lenz2025jamba} have recently emerged, combining the efficiency of state-space models with the expressivity of attention.
Inspired by these developments, we explore a hybrid architecture tailored for long video understanding that inherits the complementary strengths of both model families.

Another major bottleneck arises from the redundancy in long video sequences.
For example, a one-hour video sampled at 1 frame per second, with each frame encoded into 768 vision tokens~\cite{zhai2023siglip}, produces approximately 2.7 million tokens, exceeding even the million-token context limit of Gemini~\cite{comanici2025gemini2.5pro}.
Thus, reducing contextual length is crucial for scalable long-video modeling.
Most prior works~\cite{islam2025bimba,shen2025longvu,li2024llamavid,chen2024internvl,liu2024nvila,xu2025auroralong} address this issue by performing vision token compression and merging at the projection layer before feeding the tokens into the LLM, leveraging redundancy within ViT representations~\cite{bolya2022tokenmerge}.
However, for long videos, the LLM itself remains the primary computational bottleneck of an MLLM, as it processes the compressed sequences through billions of parameters.
Recent works have attempted to alleviate this by internal token dropping~\cite{xing2024pyramiddrop,chen2024fastv,zhang2024sparsevlm,song2024moviechat} or compression~\cite{shu2025videoxl1,ren2025vamba} within the LLM, typically guided by attention scores to identify redundant vision tokens.
Yet, developing such strategies for hybrid architectures remains largely unexplored and challenging, where the mechanism for storing token information could differ fundamentally from Transformers. 

In this work, we propose \textbf{TimeViper}, an efficient hybrid MLLM designed for long video understanding. 
Through information exchange analysis within the LLM, we identify a vision-to-text information aggregation phenomenon.  
As layer depth increases, information from vision tokens progressively converges into text tokens, across both instruction-centric tasks (e.g., video QA) and vision-centric tasks (e.g.,video captioning).
At deeper layers, even removing all vision tokens causes no performance degradation, suggesting severe token redundancy within the model.
Motivated by this observation, we introduce \textbf{TransV}, a token compression mechanism within the LLM. 
TransV progressively shortens the context length by transferring partial vision tokens into instruction tokens via gated cross-attention, preserving critical visual information while eliminating redundancy.
Extensive experiments demonstrate that TimeViper achieves promising performance to Transformer-based MLLMs across long video understanding benchmarks, including multi-choice video QA, temporal video grounding, and detailed video captioning.

Our main contributions are summarized as follows:
\begin{itemize}
    \item We introduce TimeViper, a hybrid Mamba-Transformer vision-language model for efficient long video understanding, featuring internal LLM token compression that enables processing over 10,000 frames.
    \item We discover the phenomenon of vision-to-text aggregation and vision token redundancy in hybrid architectures, and propose TransV, a mechanism that eliminates visual redundancy through explicit token information transfer.
    \item Extensive experiments demonstrate that TimeViper achieves comparable performance to Transformer-based MLLMs while accelerating inference speed.
\end{itemize}

\section{Related Works}
\label{sec:related_works}

\begin{figure*}[t]
  \centering
  \includegraphics[width=\linewidth]{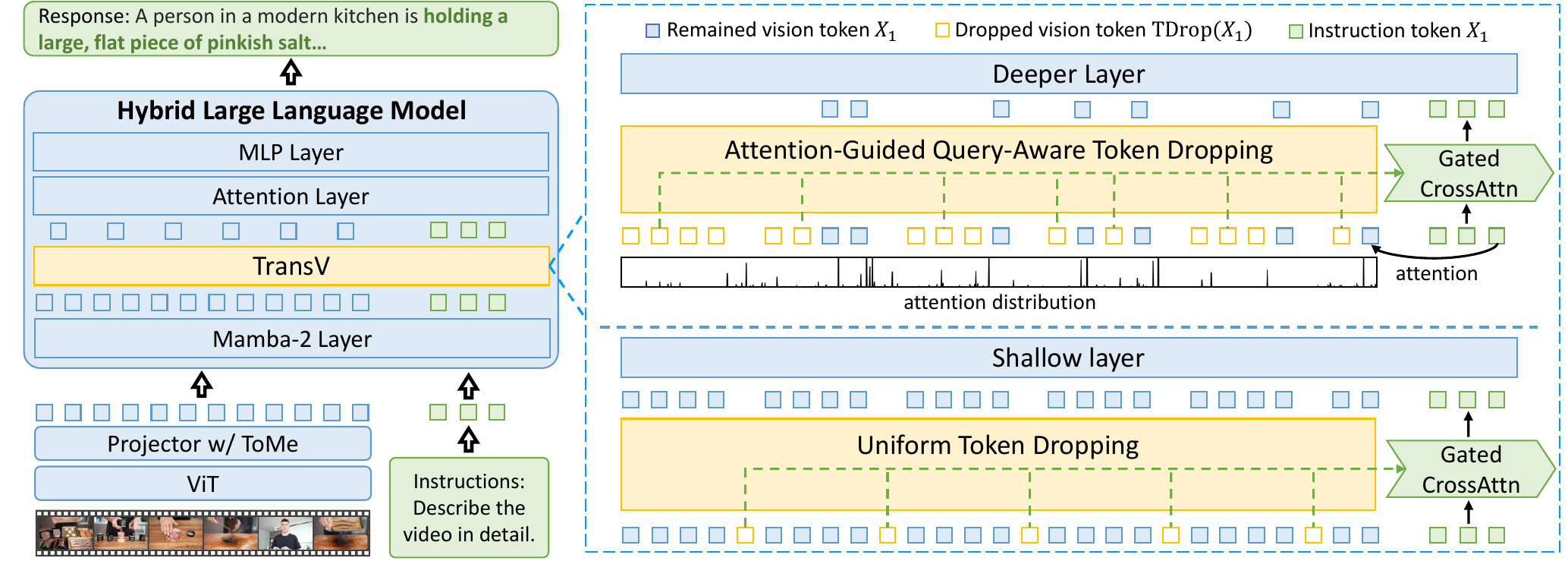}
  \caption{
Illustration of TimeViper, our proposed hybrid MLLM for long video understanding. The model consists of a ViT visual encoder, a projector with token merging, and a hybrid Mamba-Transformer LLM equipped with TransV. The token merging~\cite{bolya2022tokenmerge} compresses each frame into 16 vision tokens. Inside the LLM, TransV transfers information from redundant vision tokens to instruction tokens to reduce the number of vision tokens. Specifically, TransV uniformly drops vision tokens in shallow layers and removes low-attention vision tokens in deeper layers. The compression module is implemented through a Gated Cross-Attention mechanism~\cite{alayrac2022flamingo} with adaptive learnable weights. Note that TransV is illustrated before the attention layer for clarity, though it may be applied before any layer in practice.
  }
  \label{fig:framework}
\vspace{-8pt}
\end{figure*}

\noindent\textbf{MLLM for long video understanding.} 
Long video understanding~\cite{kahatapitiya2025langrepo,wang2024videoagent,zhang2024simpleLLM4video,mangalam2023egoschema,fu2025videomme,yue2023movie101} has long been a challenging problem in computer vision.
Towards this goal, MLLMs emerge as a promising approach, but they struggle to process long videos while comprehending content.
Existing methods aim to balance computational efficiency and performance, typically categorized into subsampling or compression strategies.
Subsampling strategies~\cite{pan2023retrieving,han2024self-retrieve,fan2024videoagent,yao2025gens,goletto2024amego,ye2025tstar} shorten video length by using language queries to retrieve the most relevant video segments.
For instance, VideoAgent~\cite{wang2024videoagent} iteratively selects frames and generates captions for them using vision-language models, which are then provided to an LLM to answer the question.
Meanwhile, compression strategies condense redundant video embeddings into more compact representations.
Most works~\cite{jin2024chatunivi,dai2023instructblip,islam2025bimba,shen2025longvu,li2024llamavid,chen2024internvl,liu2024nvila,xu2025auroralong,ren2024timechat,jiang2025storm,zohar2025apollo,li2025imove,yang2025visionzip} merge visual features before feeding them into the LLM.
For instance, LLaMA-VID~\cite{li2024llamavid} employs a dual-token strategy that compresses each frame into two tokens. 
However, these methods fail to resolve the computational bottleneck of LLM.
To further improve efficiency, another line of work drops~\cite{xing2024pyramiddrop,chen2024fastv,li2024videochatflash} or compresses~\cite{zhang2024sparsevlm,ye2025vocollama,shu2025videoxl1,ren2025vamba,alayrac2022flamingo} tokens within LLMs.
For example, PDrop~\cite{xing2024pyramiddrop} progressively prunes vision tokens across LLM layers.
Although efficient, dropping tokens based on attention scores~\cite{zou2025dont,alvar2025divprune} can cause irreversible information loss.
While token dropping is convenient and can be applied in a training-free manner, token compression avoids information loss.
VoCo-LLaMA~\cite{ye2025vocollama} and Video-XL~\cite{shu2025videoxl1} suggest compressing vision tokens into new special tokens.
Nevertheless, as existing methods rely heavily on Transformers, the token compression approaches in hybrid MLLM remain unexplored.
In this work, we pioneer this direction by introducing TimeViper, a hybrid MLLM with a token information transfer module within LLM named TransV to compress vision tokens into instruction tokens.

\noindent\textbf{State-space model for visual perception.}
Transformers’ attention with quadratic computational cost remains a fundamental bottleneck for efficiency.
Linearized architectures~\cite{peng2023rwkv,gu2024mamba} have evolved again in NLP community, aiming to reduce complexity to linear time and eliminate the need for KV-cache during inference.
Some recent efforts are pushing LLMs with linearized modules or Mamba-Transformer hybrid architectures, such as Nemotron-Nano~\cite{basant2025nanov2}, Samba~\cite{ren2025samba}, and Hymba~\cite{dong2025hymba}.
Inspired by these works, researchers have begun to explore linearized architectures for computer vision tasks like image~\cite{fan2024MALA,zhu2024visionmamba,yu2025mambaout}, video~\cite{li2024videomamba}, and 3D~\cite{liu2025map} understanding.
The rise of linearized LLM has also inspired efficient multimodal models~\cite{zhao2025cobra,xu2025auroralong,ren2025vamba,qiao2024vlmamba,liao2025mmmamba,wang2024longllava}.
However, since images and short clips involve relatively limited sequence lengths, the advantage from linearized architectures is still a controversial problem~\cite{yu2025mambaout}.
In contrast, long video understanding naturally demands models capable of processing extremely long contexts, posing far stricter efficiency requirements.
Recently, AuroraLong~\cite{xu2025auroralong} combines a ViT with RWKV6~\cite{peng2024rwkv6} and employs token merging in a projector to compress vision tokens.
In this work, TimeViper is the first hybrid MLLM that performs token compression within the hybrid LLM, achieving promising performance among 7B-sized MLLMs while maintaining high efficiency.

\section{Method}
\label{sec:method}
We propose TimeViper, a hybrid Mamba-Transformer vision-language model equipped with an internal LLM compression module, TransV, for long video understanding. 
Our method is built upon two key ideas: 
1) hybrid MLLM construction, which integrate the efficiency of state-space models with the expressivity of attention mechanisms, and 
2) performing LLM token compression through vision-to-text information transfer. 
We first introduce the hybrid model structure in~\Cref{subsec:hybrid_structure}. 
Next, in~\Cref{subsec:token_transfer}, we analyze the token information exchange between vision and text tokens and present our compression module TransV. 
Finally, we describe the training strategy in~\Cref{subsec:training}.

\subsection{Model Architecture}
\label{subsec:hybrid_structure}
Our model follows the standard multimodal design~\cite{liu2023llava} and consists of three components: a visual encoder (ViT)~\cite{zhai2023siglip}, a projector, and a hybrid Mamba-Transformer LLM~\cite{basant2025nanov2}. The LLM backbone includes 27 Mamba-2~\cite{dao2024mamba2} layers, 4 self-attention layers, and 25 MLP layers.
Following prior work~\cite{zohar2025apollo,li2024videochatflash}, we apply token merging (ToMe)~\cite{bolya2022tokenmerge} in the projection layer to reduce intra-frame redundancy. 
Given a long video with a corresponding textual instruction, the ViT encodes video frames, and the projector with ToMe produces a sequence of compressed vision tokens $X_0\in\mathbb{R}^{T_0\times D}$, while the instruction is tokenized into text instruction tokens $X_1\in\mathbb{R}^{T_1\times D}$,
where typically $T_0 \gg T_1$, and $D$ is the hidden dimension.
The LLM processes the concatenated multimodal input $X=[X_0, X_1]\in\mathbb{R}^{T\times D}$ of sequence length $T$ and generates response tokens $Y$.

The hybrid backbone integrates Mamba-2 and self-attention layers, each contributing complementary capabilities: 
the Mamba-2 layer is mainly responsible for sequence position modeling, encoding historical sequence information into a fixed-size implicit hidden memory through forgetting and memorization mechanisms, while the self-attention layer preserves the entire history of the sequence and performs retrieval and querying based on the importance of the tokens.

\noindent\textbf{Mamba-2 Layer.} 
A Mamba-2 layer is built around a core state-space model (SSM) block, which recurrently maintains a compact hidden state summarizing past information. 
Let $x_t$ denote the input at step $t$, and $h_t\in\mathbb{R}^{N\times D}$ the hidden memory. The SSM update is defined as:
\begin{equation}
    \begin{split}
        h_{t} &= A_{t} h_{t-1} + B_{t} x_{t}\ \\
        y_{t} &= C_{t}^T h_{t}\
    \end{split}
\label{eq:ssm}
\end{equation}
where $A_t$, $B_t$, and $C_t$ are discretized SSM parameters~\cite{dao2024mamba2}.
This mechanism encodes temporal dependencies via learnable decay and gating dynamics, enabling efficient information propagation over long sequences. 

\noindent\textbf{Self-Attention Layer.} 
In contrast, the self-attention layer directly models token interactions:
\begin{equation}
        y = \mathrm{SoftMax}(L\odot \frac{QK^T}{\sqrt{D}})\cdot V
\label{eq:self-attention}
\end{equation}
where $[Q,K,V]=[W_Q,W_K,W_V]X$, and $W_Q,W_K,W_V$ are learnable parameters. $L$ is the causal attention mask. 

By integrating these two mechanisms, the hybrid LLM retains the contextual expressivity of attention while benefiting from the efficiency of SSMs. 

\begin{figure}[t]
  \centering
  \includegraphics[width=1\linewidth]{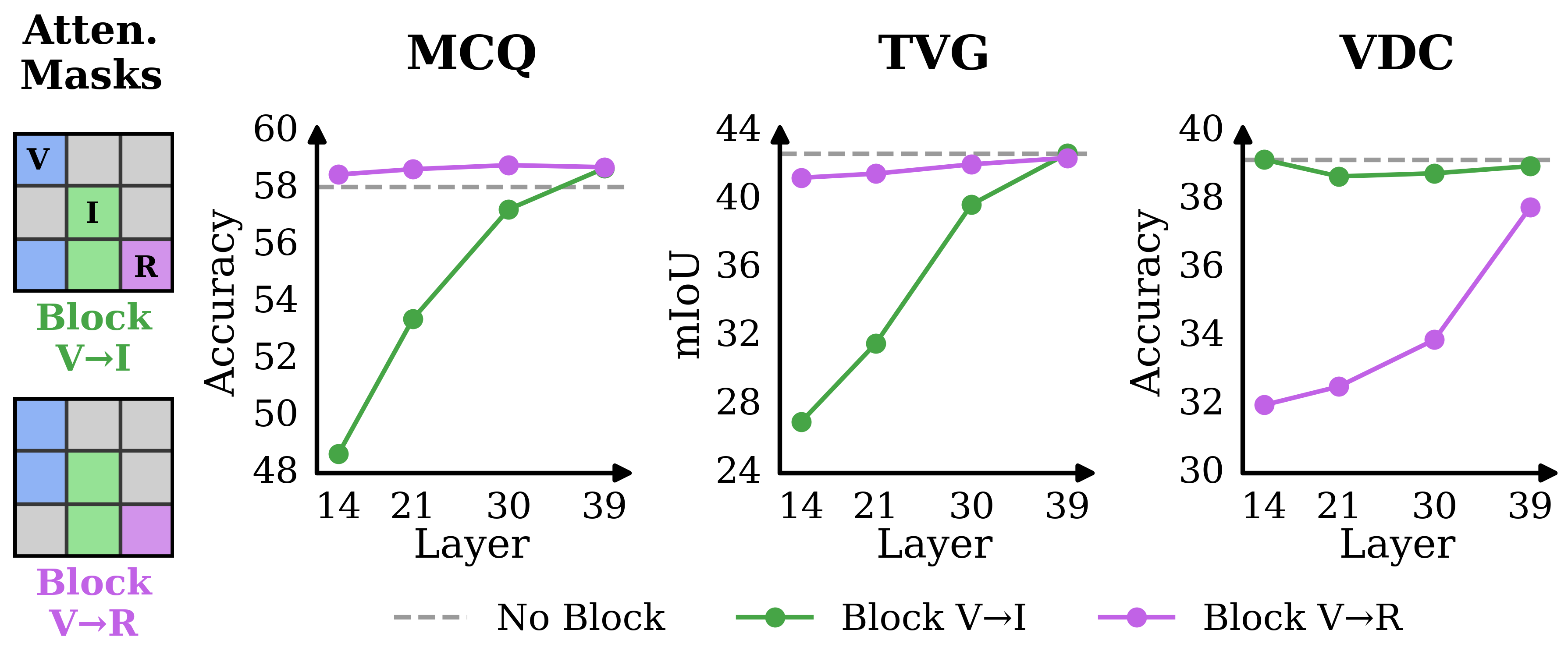}
  \caption{
Comparison of information blocking to illustrate the vision-to-text information aggregation phenomenon in hybrid MLLMs.
For instruction-centric tasks (e.g., multi-choice video QA), information is first aggregated from vision tokens to instruction tokens, which are then used for response generation.
In contrast, for vision-centric tasks (e.g., detailed video captioning), vision tokens directly contribute to response generation.}
  \label{fig:motivation_transv-token_aggregation}
  \vspace{-8pt}	

\end{figure}

\subsection{Token Information Transfer}
\label{subsec:token_transfer}
To analyze information dynamics within hybrid MLLMs, given the lack of hybrid MLLMs, we first train a hybrid model on open-source datasets for subsequent experiments and analyses.
To ensure the generalizability of our experiments on downstream tasks, we conduct the following analyses on high-quality and widely used benchmarks, including \textit{instruction-centric tasks} such as multi-choice video QA (MCQ) on VideoMME~\cite{fu2025videomme} and temporal video grounding (TVG) on Charades~\cite{sigurdsson2018charades}, and \textit{vision-centric tasks} such as video detailed captioning (VDC) on VDC~\cite{chai2025auroracap}. 
We employ standard evaluation metrics for each task, \ie, accuracy for VQA, mIoU for TVG, and LLM-judged scores for VDC.

\noindent\textbf{Vision-to-text information aggregation phenomenon.}
To understand the pattern of information flow within the hybrid model, we investigate the mechanism of information exchange among vision, instruction, and response tokens in attention layers during autoregressive generation, following the methodology  of~\cite{kaduri2025whatsintheimage}.
We apply attention masks $L$ and set the corresponding matrix values to 0 to block information exchange among different types of tokens.
For clarity, we illustrate our two information-blocking configurations, \ie, vision-to-instruction (V2I) and vision-to-response (V2R),
by assuming that $X_0$, $X_1$, and $Y$ each contain only a single token at the $l$-th layer: 

\begin{itemize}
    \item \noindent block the information from vision to instruction tokens:
\end{itemize}
\begin{equation}
    \begin{bmatrix}
    X^{l+1}_0,X^{l+1}_1,Y^{l+1}_{:t}
    \end{bmatrix} = \begin{bmatrix}
    1 & 0 & 0 \\
    0 & 1 & 0 \\
    1 & 1 & 1
    \end{bmatrix}\cdot
    \begin{bmatrix}
    X^{l}_0,X^{l}_1,Y^{l}_{:t}
    \end{bmatrix}
\label{eq:block_v2i}
\end{equation}
\begin{itemize}
    \item block the information from vision to response tokens:
\end{itemize}
\begin{equation}
\begin{bmatrix}
    X^{l+1}_0,X^{l+1}_1,Y^{l+1}_{:t}
    \end{bmatrix} = \begin{bmatrix}
1 & 0 & 0 \\
1 & 1 & 0 \\
0 & 1 & 1
\end{bmatrix}\cdot
\begin{bmatrix}
X^{l}_0,X^{l}_1,Y^{l}_{:t}
\end{bmatrix}
\label{eq:block_v2r}
\end{equation}

As shown in ~\Cref{fig:motivation_transv-token_aggregation}, we observe a consistent vision-to-text aggregation phenomenon: in instruction-centric tasks, visual information is progressively absorbed into instruction tokens until deeper layers, whereas in vision-centric tasks, vision tokens directly contribute to response generation. 
Blocking V2I drastically degrades performance in early layers for MCQ and TVG, but has negligible impact in later layers, confirming that instruction tokens eventually internalize visual cues. 
Conversely, in VDC, blocking V2R causes a sharp drop in shallow layers, highlighting the dominant role of vision tokens in direct response generation.

\begin{figure}[t]
  \centering
  \includegraphics[width=1\linewidth]{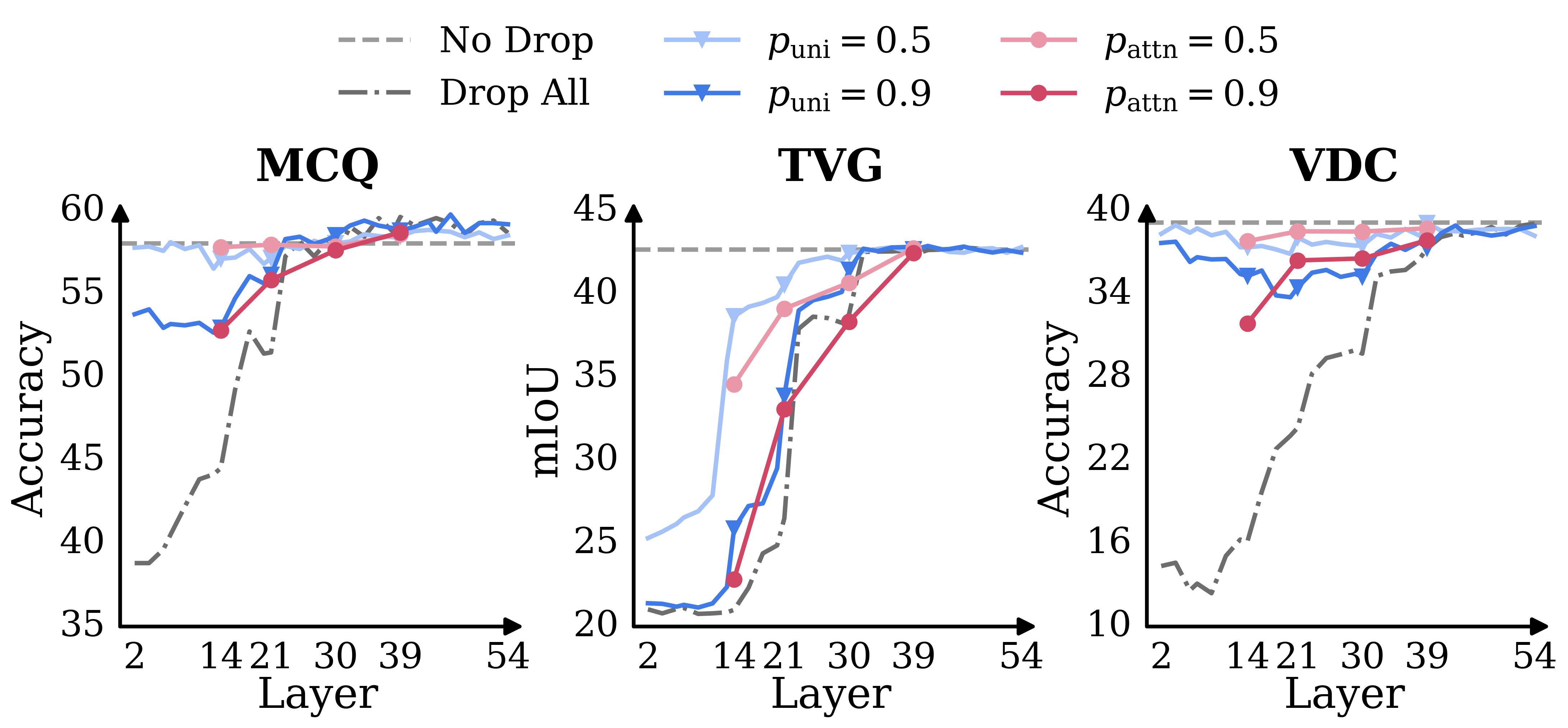}
  \caption{ 
  Illustration of token redundancy. We compare performance under different vision-token dropping rates $p$ using uniform dropping and attention-guided dropping strategies.
  In the hybrid MLLM, vision token redundancy increases progressively with layer depth, allowing more aggressive token removal in deeper layers with minimal performance loss.  
  }
  \label{fig:motivation_transv-redundancy_increase_with_layer}
\vspace{-8pt}
\end{figure}

\noindent\textbf{Vision token redundancy in hybrid MLLM.}
While many previous studies have shown vision token redundancy in Transformers~\cite{chen2024fastv,xing2024pyramiddrop,bolya2022tokenmerge}, it remains unclear how such visual redundancy manifests in hybrid models.
To investigate this, we drop vision tokens at different layers to observe performance changes on benchmarks.
Specifically, we use two vision token dropping strategies: uniform dropping (uni) and attention-guided dropping (attn), which keeps  the top-$k$ vision tokens most attended by the last instruction token $X_{T_1}$. 
Let $p$ denote the token dropping rate, $T_d = p T_0$ be the number of tokens to be dropped.
We define the dropping operator $\mathrm{TD}(\cdot)$ as:
\begin{equation}
    \mathrm{TD}(X)=
    \begin{cases}
        \mathrm{Uniform}(X, T_d) &\text{uni} \\
        {{\mathrm{Topk}}(X,-\text{Attn}(X_{T_1},X),T_d}) &\text{attn}
    \end{cases}
\end{equation}
where $\mathrm{Uniform}(X,k)$ uniformly drops $k$ tokens from $X \in \mathbb{R}^{N}$, $\mathrm{TopK}(X,S,k)$ discards $k$ tokens with the highest scores in $X$ according to $S \in \mathbb{R}^{N}$, and $\mathrm{Attn}(\cdot, \cdot)$ computes the attention scores using the first argument as the query and the second as the key.
Results in~\Cref{fig:motivation_transv-redundancy_increase_with_layer} show that redundancy increases with depth.
For MCQ and VDC, tokens can be uniformly dropped at all layers, but attention-guided dropping is reliable only in deeper layers. 
In TVG, excessive token dropping before the first attention layer, \ie, the 14-th layer, harms performance, but the drop ratio can increase in later layers.
Vision tokens are critical in shallow layers but become nearly 100\% redundant in deep layers across our testbeds.
For all tasks, even discarding all vision tokens in deep layers, the model can still achieve high performance by relying solely on the instruction tokens, which is surprisingly similar to observations from previous Transformer-based image MLLMs~\cite{chen2024fastv,zhang2025llavamini,zhang2025cross}.

\noindent\textbf{Token information transfer via TransV.}
Motivated by these findings that information implicitly transfers from vision to text tokens and vision token redundancy is severe for all tasks, we propose TransV, a lightweight in-LLM compression module that explicitly transfers visual information into instruction tokens, reducing redundant computation while preserving task performance. 
At the $l$-th layer, the token information transfer from vision to instruction tokens is formulated as:
\begin{equation}
    \begin{split}
        \tilde{X}_1^l &= \mathrm{CrossAttn}_l(X_1^l, \mathrm{TD}_l(X_0^l))    \\
        X_1^{l+1} &= X_1^l + \tanh(\alpha_l)(\tilde{X}_1^l)
    \end{split}
\end{equation}
where the $\mathrm{CrossAttn}(\cdot,\cdot)$ computes the attention with the first term as the query and the second as both the key and value. 
$\alpha_l$ is a learnable scalar controlling the degree of information aggregation, and its value is normalized to the range $[-1,1]$ via $\tanh(\cdot)$. 
$\alpha_l$ is initialized to zero to ensure instruction understanding.

\subsection{Training Procedure}
\label{subsec:training}
To effectively adapt TimeViper for long video understanding, we divide the training process into two stages and train the model using fully open-source data:
(1) Image-text alignment stage: 
We first pretrain the projector to align the ViT and LLM modalities using 3M high-quality image–text pairs sampled from  CC12M~\cite{changpinyo2021cc12m} and PixelProse~\cite{singla2024pixelprose}. Token compression is disabled during this stage. 
(2) Visual instruction tuning: 
We then fine-tune the projector and LLM, including the compression modules, on approximately 4.8M multimodal instruction pairs, consisting of 1.8M video instruction data~\cite{sharegemini,bain2021webvid,carreira2017k400,zhang2025llavavideo,chen2024sharegpt4video,li2024videochatflash,song2024moviechat} primarily sourced from LLaVA-Video, 
2.8M single-image instruction data~\cite{lillavaov} from LLaVA-OneVision, 
and diverse downstream task-specific datasets including 
26K dense video captioning samples~\cite{caba2015activitynet,zhou2018youcook2,tang2019coin,zala2023hirest,huang2020vitt} and 
250K temporal video grounding samples~\cite{liu2024etbench,wang2025timer1,yang2023vid2seq,anne2017didemo,afouras2023htstep,miech2019howto100m,zeng2025timesuite,ren2024timechat,guo2025vtgllm,oncescu2021queryd,wang2024internvid}.
This stage adapts TimeViper for instruction-following and video understanding while learning effective internal compression through TransV.



\section{Experiments}
\label{sec:exp}
We first describe the experimental setups in~\Cref{subsec: exp_setups}, followed by the ablation studies in~\Cref{subsec: abl_studies} and main results in~\Cref{subsec: main_results}. 
Finally, in~\Cref{subsec: analyses}, we provide a qualitative analysis illustrating how hybrid models and Transformers interpret visual content through attention visualizations.

\subsection{Experimental Setup}
\label{subsec: exp_setups}

\noindent\textbf{Downstream benchmarks.}
We evaluate TimeViper across a diverse suite of video understanding benchmarks:
(1) \textit{VideoMME}~\cite{fu2025videomme}: A comprehensive video QA benchmark covering multiple domains. It includes 2.7K QA samples over videos ranging from 11 seconds to 1 hour. We evaluate models without textual subtitles.
(2) \textit{LVBench}~\cite{wang2025lvbench}: A benchmark on hour-long video understanding across six dimensions, comprising 2094 multiple-choice QA samples.
(3) \textit{MLVU}~\cite{zhou2025mlvu}: Designed for minute-level video understanding, with 2174 QA samples spanning diverse domains.
We evaluate the average performance of multiple-choice tasks (M-Avg), where videos have an average duration of 653 seconds.
(4) \textit{LongVideoBench}~\cite{wu2024longvideobench}: Targets long-form referring reasoning that requires retrieval-based QA, with videos averaging 473 seconds.
(5) \textit{MVBench}~\cite{li2024mvbench}: A short-term video QA benchmark emphasizing temporal reasoning, containing 4K QA pairs over 20 task categories. 
(6) \textit{Charades}~\cite{sigurdsson2018charades}: A temporal video grounding benchmark containing 6672 minute-level indoor activity videos paired with natural language queries.
(7) \textit{VDC}~\cite{chai2025auroracap}: An efficient and high-fidelity video captioning benchmark evaluated via query-conditioned scoring using LLaMA3-8B~\cite{dubey2024llama3}. It consists of 1027 videos and we evaluate on the ``detailed'' split.

\begin{table}[t]
\centering
\caption{Ablation of TransV choices.
The ``uni\_7\_0.5-attn\_39\_0.9'' denotes applying uniform TranV at the 7th layer with a dropping rate of $p=50\%$ and attention-guided TransV at the 39th layer with $p=90\%$.
``TDuni'' denotes uniform token dropping. }
\scalebox{0.7}{
\begin{tabular}{c|lc|cccc}
\toprule
& Method   & max frame    & VideoMME & VDC & Charades  \\
\midrule
1 & none    & 5k    &  58.8 & 39.8 & 40.5 \\
\midrule

2 & TDuni\_7\_0.5                & 8k     & 57.3    & 39.0    & 26.1 \\

3 & uni\_7\_0.5& 8k  & 56.7 & 38.9 & 38.1 \\
4 & uni\_2\_0.5& 9k  & 56.1  & 39.7 & 38.2 \\

5 &uni\_7\_0.9 &  $>$10k &  53.4 & 37.9 & 34.6 \\
\midrule
6 & uni\_7\_0.5-uni\_39\_0.9     & $>$10k & 56.2 & 39.4 & 37.9 \\
7 & uni\_7\_0.5-attn\_39\_0.9    & $>$10k & 56.6 & 39.1 & 37.9 \\
\bottomrule
\end{tabular}
}
\label{tab:abl_TransV_module}
\end{table}
\begin{figure}[t]
  \centering
  \includegraphics[width=0.9\linewidth]{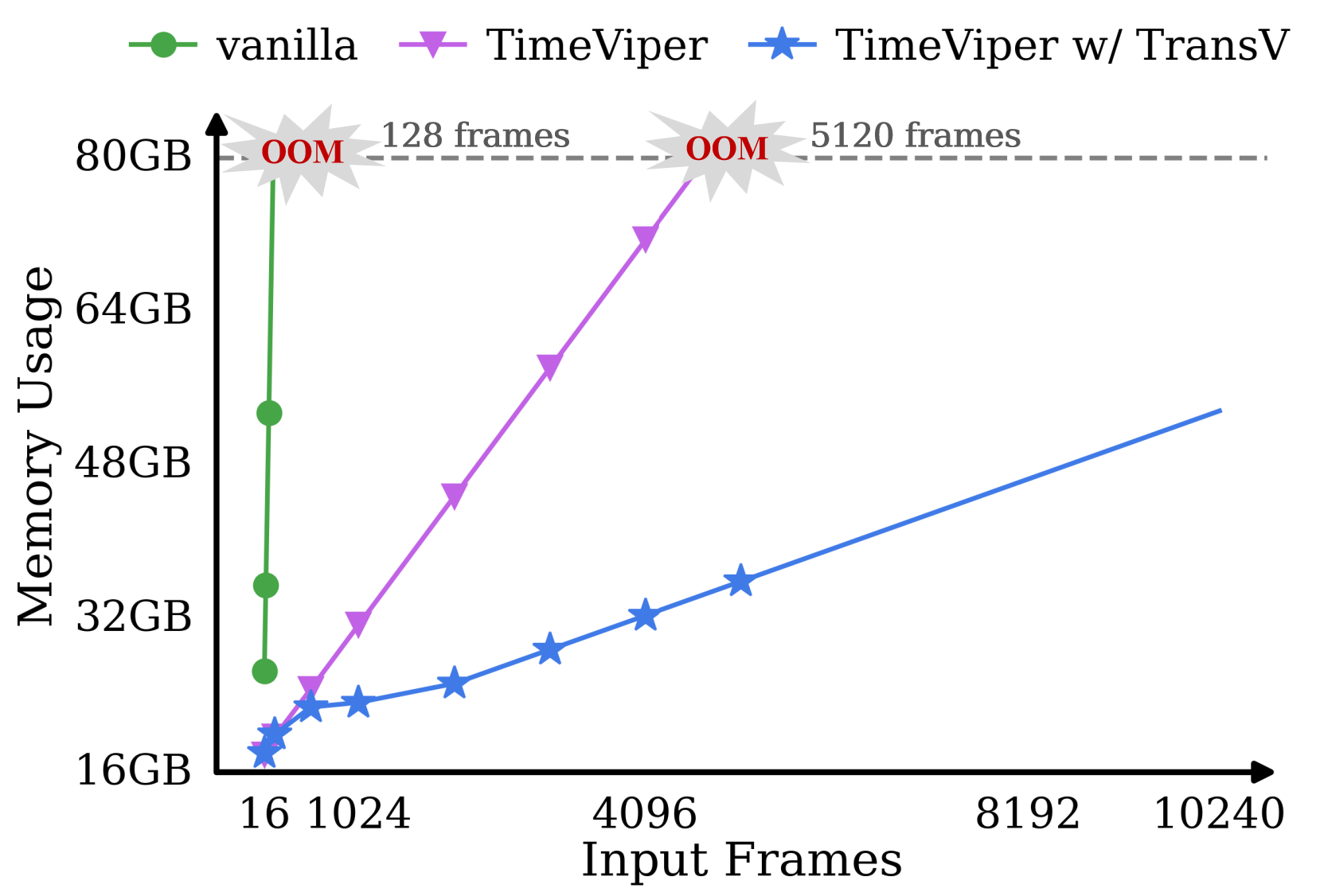}
  \caption{Comparison of GPU memory usage during inference.
  While ToMe extends the context window to about 5K frames, TransV efficiently scales beyond 10K frames.
  }
  \label{fig: frames_GPUMem}
\vspace{-8pt}
\end{figure}

\noindent\textbf{Implementation details.}
Our data are organized in the order of system prompt tokens, video tokens, and instruction token as the LLM inputs.
For all training and evaluation processes, videos are sampled at 1 frame per second.
During training, videos longer than 256 frames are uniformly sampled to 256 frames;
during evaluation, we use at most the first 256 frames.
Each input frame is resized to 384×384 and initially encoded into 768 vision tokens. 
After being projected with ToMe, each frame is compressed into 16 tokens~\cite{zohar2025apollo,li2024videochatflash}.
We apply TransV at the 7th shallow LLM layer with token dropping rate $p=50\%$, and the 39th deep LLM layer where TransV is applied using an attention-guided strategy with $p=90\%$.
Introducing TransV adds approximately 100M parameters to the model. 
To accelerate training, we implement training with data packing~\cite{chen2025eagle2.5} that supports training with varied sequence length caused by TransV.
Across all training stages, the model uses a learning rate of 1e-5, AdamW optimizer with weight decay of 0.01, warmup rate of 0.03, and cosine annealing scheduler.
For TransV modules, we adopt a higher learning rate of 5e-5.

\begin{figure}[t]
  \centering
  \includegraphics[width=0.9\linewidth]{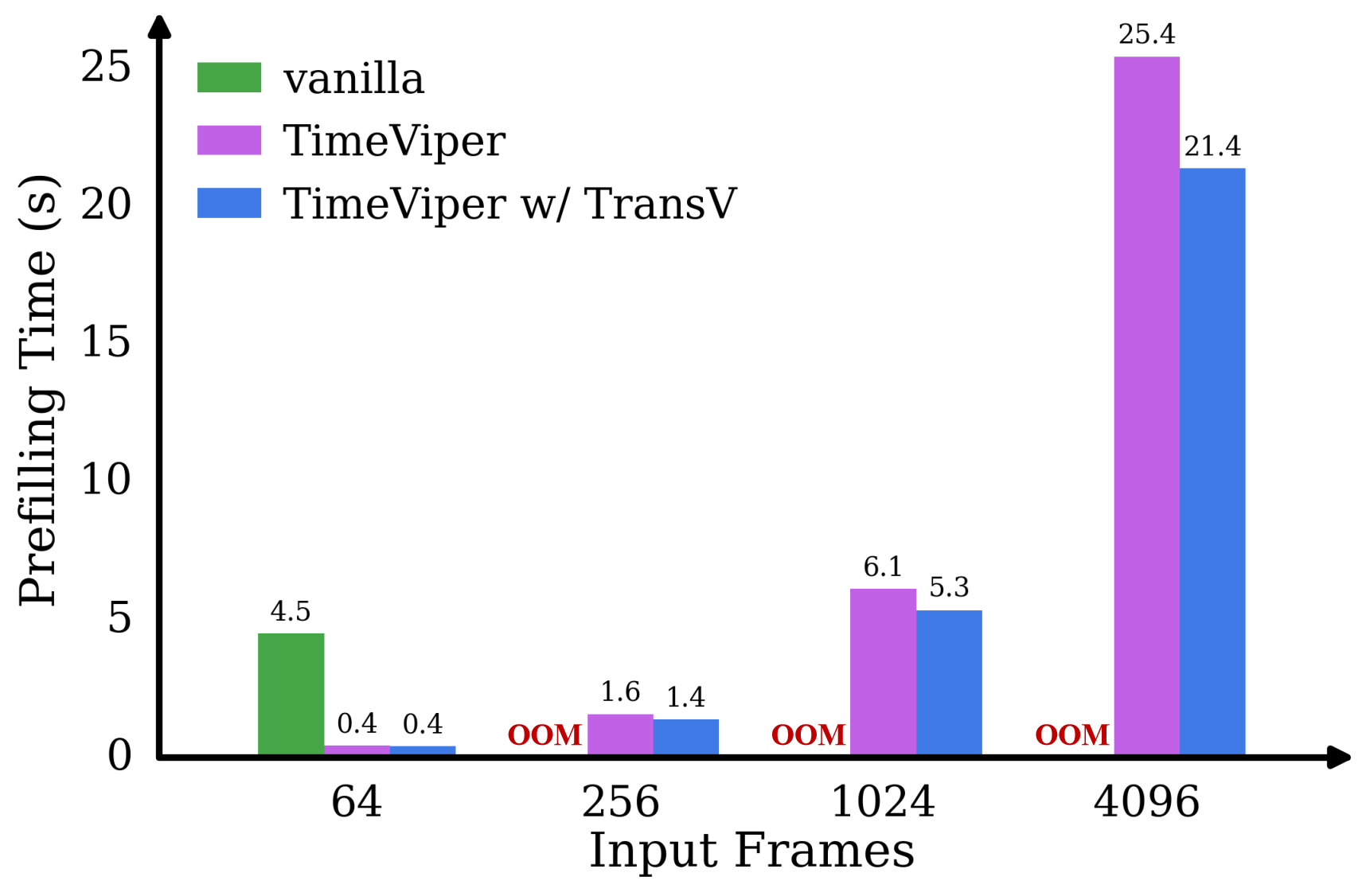}
  \caption{ 
  Comparison of prefilling time.
  TransV incurs no additional latency at low frame inputs (e.g., 64 frames) while significantly reducing prefilling time at high frame inputs. 
  For instance, at 4,096 frames, TransV reduces prefilling time by 15.7\% compared to the ToMe baseline.
  }
  \label{fig: frames_prefilling_time}
\vspace{-8pt}
\end{figure}

\subsection{Ablation Study}
\label{subsec: abl_studies}
For representation simplicity, each TransV is denoted as \textit{``type\_layer\_ratio-...''}, for example, \textit{``uni\_7\_0.5-attn\_39\_0.9''} represents applying uniform TransV in the 7th layer with a ratio of 50\% and attention-guided TransV in the 39th layer with ratio of 90\%.

\noindent\textbf{Impact of compression components on GPU memory consumption.}
We apply ToMe in the projector and TransV in the LLM. Benefiting from the hybrid Mamba-Transformer backbone, both memory usage and prefilling time generally grow approximately linearly with input length.
\Cref{fig: frames_GPUMem} reports memory usage as the number of frames increases. The vanilla model runs ``Out of Memory'' error at merely 128 frames. 
TimeViper applying ToMe in the projector alleviates the initial token burden, extending the limit to approximately 5K frames. 
TimeViper with TransV further enables better scalability: at 4,096 frames, it reduces memory consumption by 54.8\% compared to TimeViper, and can handle 10K+ frames with ample margin.
This highlights the complementary roles of token compression in the projector and within the LLM.

\noindent\textbf{Impact of compression components on prefilling time.}
As shown in~\Cref{fig: frames_prefilling_time},the vanilla model already incurs 4.5s latency at 64 frames. 
TimeViper drastically reduces this to 0.4s, and TransV further decreases prefilling time, with the effect becoming more pronounced as the number of frames increases. 
Notably, at 4,096 frames, TransV reduces prefilling time by 15.7\% compared to TimeViper.

\begin{table*}[t]
\centering
\caption{Comparison with state-of-the-art models.
Our work differs from previous studies both the choice of LLM backbone and the design of token compression strategy, while achieving competitive performance across benchmarks. 
Most existing methods fine-tune the ViT (indicated with *), whereas we do not due to computational constraints. 
Additionally, while the concurrent work Nanov2-VL~\cite{deshmukh2025nanov2vl} is trained on 46.7M samples, we uses only 7.8M, making Nanov2-VL a reasonable upper bound for hybrid models.
}
\label{tab:comp_sota}
\resizebox{\textwidth}{!}{%
\begin{tabular}{l c c c c c cc c c c}
\toprule
\multirow{2}{*}{\textbf{Model}} & 
\multirow{2}{*}{\textbf{LLM}} & 
\multirow{2}{*}{\textbf{$>$10K frame input}} & 
\textbf{MVBench} & 
\textbf{LongVideoBench} & 
\textbf{MLVU} & 
\multicolumn{2}{c}{\textbf{VideoMME}} & 
\textbf{LVBench} & 
\textbf{Charades-STA} & 
\textbf{VDC} \\
\cmidrule(lr){7-8}
& &  & avg.acc & val & M-Avg & overall & long & avg.acc & mIoU & avg.acc \\
\midrule
\multicolumn{11}{c}{Proprietary Models} \\
\midrule
GPT-4V~\cite{achiam2023gpt4v} & - & - & 43.7 & 59.1 & 49.2 & 59.9 & 53.5 & - & - & - \\
GPT-4o~\cite{hurst2024gpto} & - & - & 64.6 & 66.7 & 64.6 & 71.9 & 65.3 & 30.8 & 35.7 & - \\
Gemini-1.5-Pro~\cite{team2024gemini1.5pro} & - & - & 60.5 & 64.0 & - & 75.0 & 67.4 & 33.1 & - & 43.1 \\
\midrule
\multicolumn{11}{c}{Transformer-based Video MLLMs} \\
\midrule
LLaMA-VID~\cite{li2024llamavid} & Vicuna-1.5-7B~\cite{touvron2023llama2} & \ding{55} & 41.9 & - & 33.2 & 25.9 & - & 23.9 & - & 25.6 \\ 
LongVA~\cite{zhang2024longva} &  Qwen2-7B~\cite{team2024qwen2} & \ding{55} & - & - & 56.3 & 52.6 & 46.2 & - & - & 27.9 \\
LongVU~\cite{shen2025longvu} & Qwen2-7B~\cite{team2024qwen2} & \ding{55} & 66.9 & - & 65.4 & 60.6 & 59.5 & - & - & - \\ 
VILA1.5-7B~\cite{fu2025vita1.5} & Qwen2-7B~\cite{team2024qwen2} & \ding{55} & 56.8 & - & 56.8 & 58.8 & - & - & - & - \\
LLaVA-OneVision*~\cite{lillavaov} & Qwen2-7B~\cite{team2024qwen2} & \ding{55} & 56.7 & 56.3 & 64.7 & 58.2 & - & - & 13.5 & 41.2 \\
LLaVA-Video*~\cite{zhang2025llavavideo} & Qwen2-7B~\cite{team2024qwen2} & \ding{55} & 58.6 & 58.2 & 70.8 & 63.3 & - & - & - & - \\
Qwen2-7B-VL*~\cite{wang2024qwen2vl} & Qwen2-7B~\cite{team2024qwen2} & \ding{55} & 67.0 & - & - & 63.3 & - & - & - & 41.6 \\ 
Qwen2.5-VL*~\cite{bai2025qwen2.5vl} & Qwen2.5-7B~\cite{qwen2.5} & \ding{55} & 69.6 & 56.0 & 70.2 & 65.1 & - & 45.3 & 43.6 & - \\ 
LongVILA*~\cite{longvila} & Qwen2-7B~\cite{team2024qwen2} & \ding{55} & 67.1 & 57.1 & - & 60.1 & 47.0 & - & - & - \\
Kangaroo*~\cite{kangaroogroup} & LLaMA3-8B~\cite{dubey2024llama3} & \ding{55} & 61.0 & 54.8 & 61.0 & 56.0 & 46.7 & 39.4 & - & - \\
Video-XL*~\cite{shu2025videoxl1} & Qwen2-7B~\cite{team2024qwen2} & \ding{55} & 55.3 & 50.7 & 64.9 & 55.5 & - & - & - & - \\ 
Vamba*~\cite{ren2025vamba} & Qwen2-VL-7B~\cite{team2024qwen2} & \ding{55} & 60.4 & 55.9 & 65.9 & 57.8 & - & - & - & - \\ 
VideoChat-Flash*~\cite{li2024videochatflash} & Qwen2-7B~\cite{team2024qwen2} & \textcolor{blue}{\ding{51}} & 73.2 & 64.2 & 74.5 & 64.0 & 53.6 & 47.2 & 48.4 & - \\ 
VTimeLLM~\cite{shu2025videoxl1} & Vicuna-1.5-13B~\cite{touvron2023llama2} & \ding{55} & - & - & - & - & - & - & 34.6 & - \\
AuroraCap*~\cite{shu2025videoxl1} &  Vicuna-1.5-7B~\cite{touvron2023llama2} & \ding{55}  & - & - & - & - & - & - & - & 39.0 \\ 
\rowcolor[HTML]{DEEBF7}
\textbf{Qwen2.5-7B (ours)} & Qwen2.5-7B~\cite{qwen2.5} & \ding{55}  & 57.6 & 55.4 & 64.9 & 56.6 & 48.7 & 36.6 & 40.8 & 42.0 \\ 
\midrule
\multicolumn{11}{c}{Linearized/Hybrid Video MLLMs} \\
\midrule
LongLLaVA*~\cite{wang2024longllava} & Jamba-52B~\cite{lenz2025jamba} & \ding{55}  & 64.6 & 53.5 & - & 53.8 & 46.4 & - & - & - \\ 
AuroraLong*~\cite{xu2025auroralong} & RWKV6-2B~\cite{peng2024rwkv6} & \textcolor{blue}{\ding{51}}  & 53.2 & - & 52.7 & - & - & - & - & 42.5 \\ 
\textcolor{gray}{Nanov2-VL* (upper bound)~\cite{deshmukh2025nanov2vl}} & \textcolor{gray}{Nanov2-12B~\cite{basant2025nanov2}} & \textcolor{gray}{\ding{55}}  & \textcolor{gray}{-} & \textcolor{gray}{63.6} & \textcolor{gray}{73.6} & \textcolor{gray}{66.0} & \textcolor{gray}{-} & \textcolor{gray}{-} & \textcolor{gray}{-} & \textcolor{gray}{-} \\ 
\rowcolor[HTML]{DEEBF7}
\textbf{TimeViper (ours)} & Nanov2-9B~\cite{basant2025nanov2} & \ding{55} & 57.2 & 54.1 & 65.6 & 58.8 & 48.8 & 35.5 & 40.5 & 39.7 \\ 
\rowcolor[HTML]{DEEBF7}
\textbf{TimeViper (ours w/ TransV)} & Nanov2-9B~\cite{basant2025nanov2} & \textcolor{blue}{\ding{51}} & 56.2 & 52.0 & 63.1 & 56.9 & 48.2 & 35.6 & 37.9 & 39.1 \\ 
\bottomrule
\end{tabular}%
}
\end{table*}

\begin{figure*}[t]
  \centering  \includegraphics[width=0.97\linewidth]{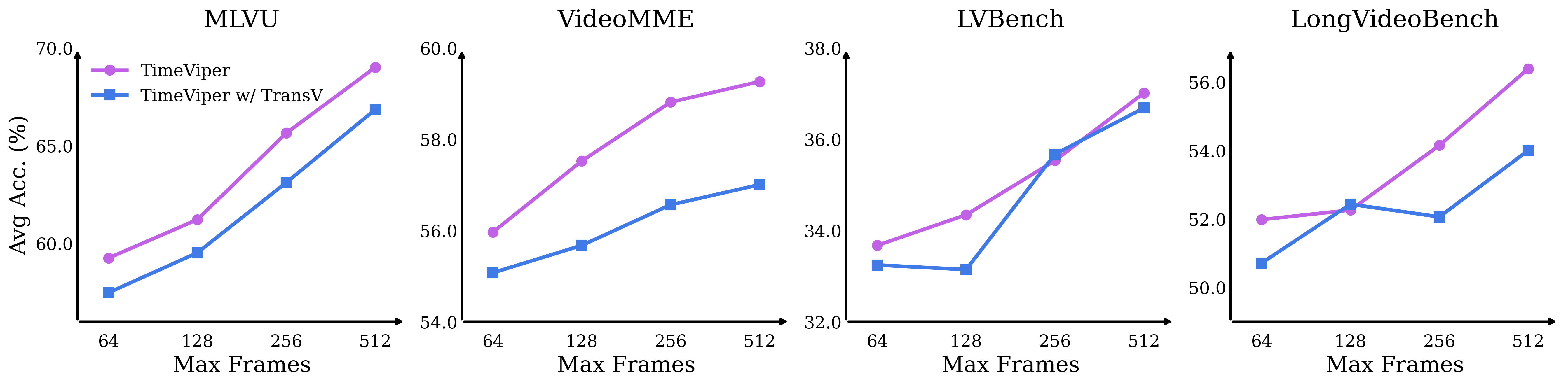}
  \caption{
  Comparison of performance as the number of input frames increases on long-video understanding benchmarks.
  We train our models with 256 frames as inputs, and sample 1 frame per second during evaluation. 
  The x-axis here denotes the maximum number of frames. 
  If a video exceeds this length, we take only the first max frames for inference.
  }
  \label{fig: Performance-framenum}
\vspace{-8pt}
\end{figure*}

\noindent\textbf{TransV placement in shallow layers.}
As shown in~\Cref{tab:abl_TransV_module}, the TimeViper baseline can process approximately 5K input frames. 
By comparing rows 1, 2, and 3, we observe that introducing token dropping or token compression enables the model to handle over 8K frames.
Comparing rows 2 and 3, using TransV effectively mitigates the Charades performance drop, from 26.1 to 38.1, indicating that it successfully facilitates token transfer.
Comparing rows 3 and 4, compressing at the 7th layer does not necessarily outperform compression at the 2nd layer on the VDC or TVG benchmarks.
For example, compression at the 7th layer outperforms the 2nd layer by 0.6 points on MCQ, but performs 0.8 points worse on VDC.

\noindent\textbf{TransV placement in deep layers.}
From rows 6 and 7 in~\Cref{tab:abl_TransV_module}, attention-guided TransV yields higher MCQ performance of 56.6 than uniform TransV's 56.2, with minor differences on VDC and Charades. 
Moreover, transferring token information in deeper layers significantly increases the model’s long-context capacity: comparing rows 1 and 3, the model handles tens of thousands of frames with only a 0.1 drop on VideoMME.

\noindent\textbf{Compression rate for TransV in shallow layers.}
We evaluate a higher compression rate of $p=90\%$ at the 7th layer in~\Cref{tab:abl_TransV_module}.
Larger compression rate allows the model to process more frames, but it comes with a significant performance drop. 
Comparing rows 4 and 5, after increasing compression rate from $50\%$ to $90\%$, the accuracy on VideoMME decreases from 56.7 to 53.4.

\begin{figure*}[t]
  \centering  \includegraphics[width=\linewidth]{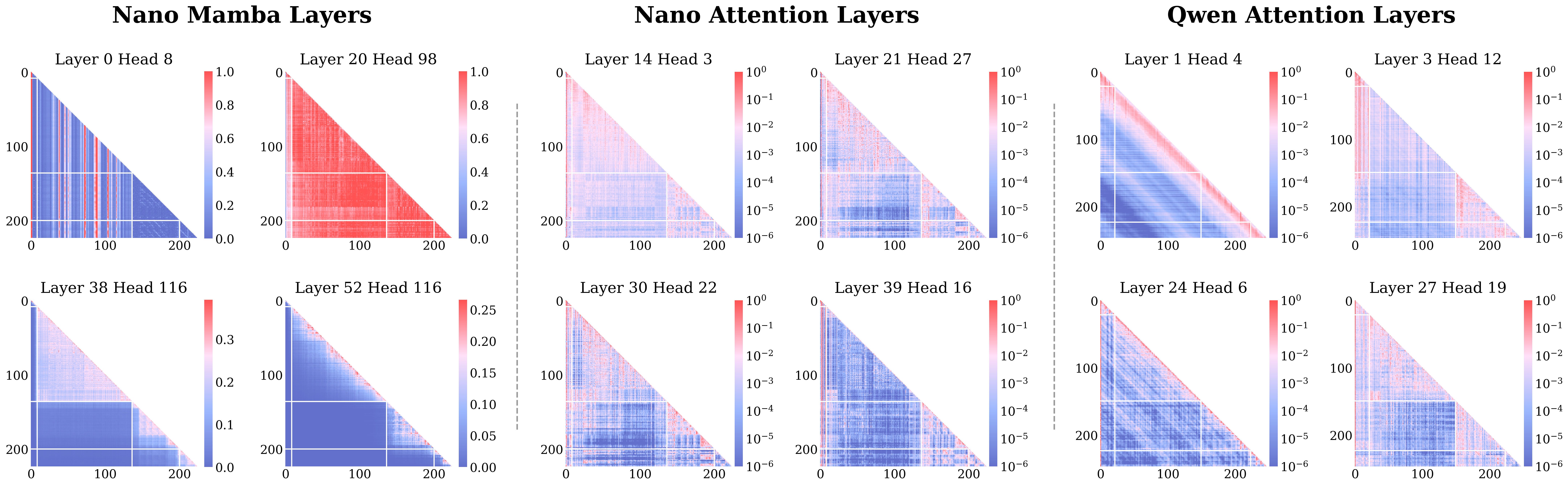}
  \caption{
      Illustration of attention score matrices in Nanov2~\cite{basant2025nanov2} and Qwen2.5~\cite{qwen2.5} at shallow and deep layers. 
      White lines divide the input sequence into four distinct segments: system prompt, vision tokens, user instruction, and the generated response.
  }
  \label{fig:exp_analyses-attn_mamba_attn_pattern}
\vspace{-8pt}	
\end{figure*}
\noindent\textbf{Impact of different LLM backbones under identical training recipe.}
To isolate the effect of model architecture from scaling up, we train a Transformer-based baseline using Qwen2.5 following exactly the same training recipe as our hybrid model. As shown in~\Cref{tab:comp_sota}, Qwen2.5 performs on par with TimeViper when trained on our dataset, suggesting that purely Transformer-based architectures do not offer a clear advantage under comparable training conditions.
Notably, the concurrent work Nanov2-VL, which adopts a standard MLLM architecture but is trained on 46.7M samples, substantially larger than our 7.8M, achieves state-of-the-art performance. 
This indicates that hybrid MLLMs can benefit significantly from scaling.

\subsection{Main Results}
\label{subsec: main_results}

\noindent\textbf{TimeViper achieves competitive performance with current models across video understanding benchmarks.}
\textit{For MCQ tasks}, 
as shown in~\Cref{tab:comp_sota}, despite not fine-tuning ViT, TimeViper with TransV achieves an average accuracy of 56.2 on VideoMME, +0.7 points higher than Video-XL (55.5), which compresses tokens into new ones within Qwen2. 
\textit{For VDC task}, TimeViper achieves strong performance with an accuracy of 39.7, excqeeding the task-specific model Auroracap by +0.7 points.
\textit{For TVG task,} 
TimeViper establishes a surprisingly strong baseline with an mIoU of 40.5 on Charades, significantly outperforming the task-specific model VTimeLLM-13B with an mIoU of 34.6. 
This is particularly notable because TimeViper uses only SigLIP positional embedding for vision tokens and relies on the implicit temporal modeling of Mamba layers. 
Yet the model learns robust temporal alignments between videos and language query, matching or exceeding prior models such as Qwen2.5-VL-7B that explicitly employ MRoPE for fine-grained timestamp modeling.
These results collectively demonstrate that hybrid Mamba-Transformer architectures are highly competitive for long video understanding.

\noindent\textbf{Effect of increasing the number of inference frames.}
Since the model is trained with 256 frames, we evaluate test-time scalability by varying the number of input frames. 
As shown in~\Cref{fig: Performance-framenum}, TimeViper scales robustly with longer contexts across four long video understanding benchmarks.
For example, when increasing the input frames from 256 to 512 frames, MLVU improves from 65.64 to 69.00, and LVBench increases from 35.53 to 37.0.

\subsection{Qualitative Analysis}
\label{subsec: analyses}
To better understand how hybrid MLLMs differ from Transformer-based MLLMs in processing multimodal inputs, we first formalize the definitions of attention scores used in both Mamba-2 and self-attention layers and then analyze attention behaviors across layers. 
For Mamba layers, we follow~\cite{ameen2025mambavis} to define the attention pattern, while for Transformer layers, we use the attention weights. 

\begin{figure*}[t]
  \centering
\includegraphics[width=\linewidth]{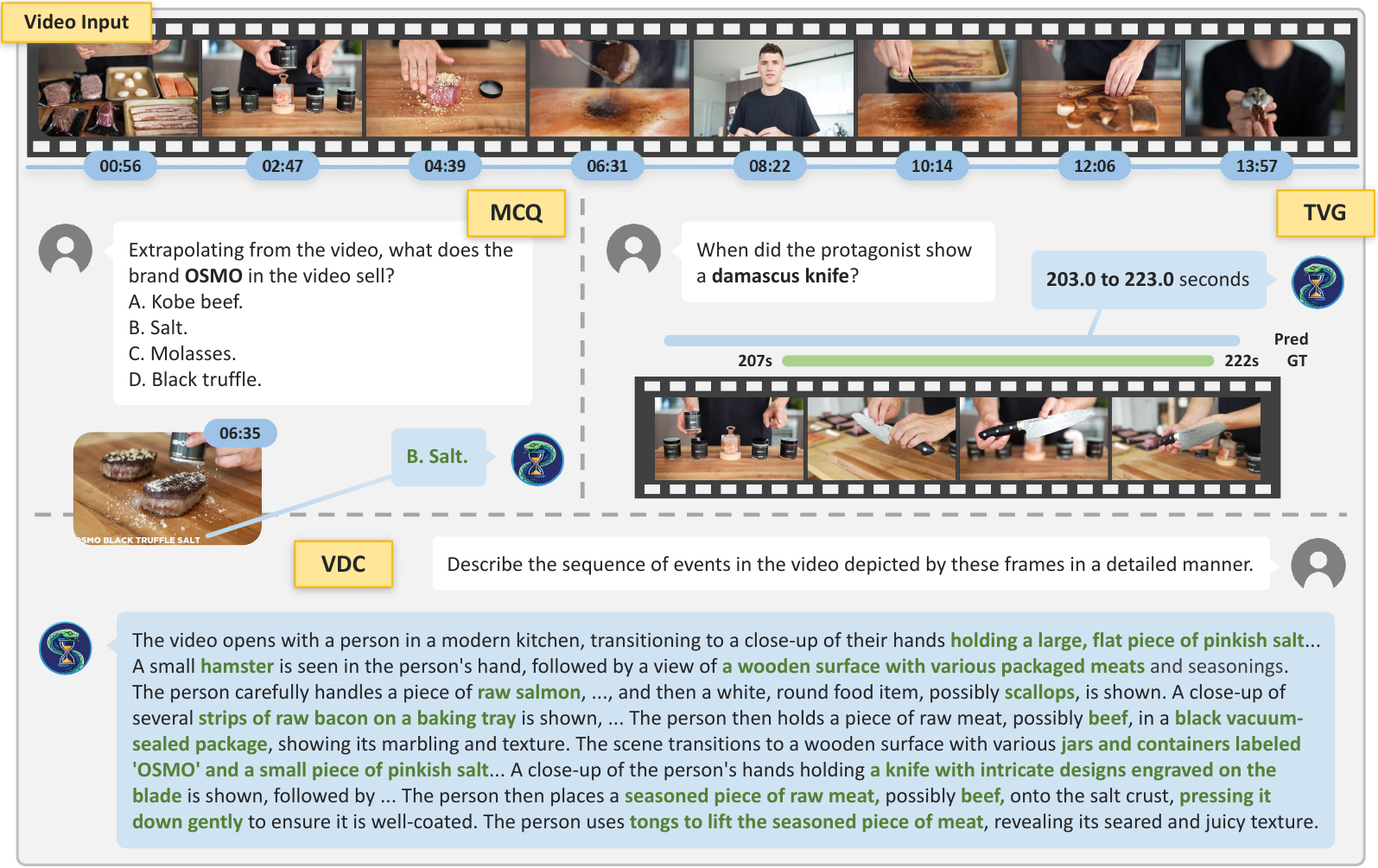}
  \caption{Qualitative results of TimeViper on three long video understanding tasks. (1) MCQ: The model demonstrates reasoning capability by correctly answering a multi-choice question about the video's content. (2) TVG: It accurately localizes the temporal boundaries for a specific event, reaching an IoU of 0.75. (3) VDC: The model generates a detailed description that showcases its fine-grained comprehension. Green text highlights accurate detailed descriptions. Some output in the middle is omitted for brevity.}
  \label{fig:case}
\vspace{-8pt}
\end{figure*}


\noindent\textbf{Definition of attention score.}
For self-attention (\Cref{eq:self-attention}), the attention score $M_{j,i} \in \mathbb{R}$ from $x_j$ to $x_i$ is:
\begin{equation}
        y_i = \mathrm{Softmax}(\frac{Q_iK_{\leq i}^T}{\sqrt{D}})\cdot V_{\leq i}=\sum_{j=1}^iM_{i,j} V_j
\end{equation}
For the SSM block, we rewrite~\Cref{eq:ssm} to express its attention pattern as the weighted sum from inputs $[x_1, \dots, x_i]$ to the output $y_i$:
\begin{equation}
    y_{i} = \sum_{j=1}^i C_{i}^T (\prod_{k=j+1}^i A_{k})B_{j} x_{j} = \sum_{j=1}^iM'_{i,j} x_{j}
\end{equation}
Here, $|M'_{i,j}| \in \mathbb{R}^+$ serves as the “attention score”~\cite{ameen2025mambavis, zimerman2025mambavis} from $x_j$ to $x_i$ within the SSM block.
Although both the self-attention and Mamba mechanisms employ a multi-head design~\cite{vaswani2017attention,dao2024mamba2} along the hidden dimension, we omit this detail in the equations for simplicity.

\noindent\textbf{Diverse and specialized attention patterns in Mamba layers.} 
~\Cref{fig:exp_analyses-attn_mamba_attn_pattern} visualizes the attention score matrices of Nano (hybrid) and Qwen (Transformer) from shallow to deep layers.
Mamba layers exhibit diverse attention patterns, ranging from sparsity, locality to globality, suggesting that different layers and heads specialize in modeling different types of dependencies. 
For example, Layer 0, Head 8 of Mamba layer (ML0) exhibits sparsity, where only a few tokens receive dominant attention from the following tokens, revealing Mamba's capability to selectively highlight salient tokens.
In contrast, ML20 exhibits globality, where all tokens attend uniformly to preceding tokens, reflecting its effective integration of prior information.
ML52 displays locality, focusing primarily on neighboring tokens.
This diversity highlights the complementary strengths of state-space modeling within hybrid architectures.

\noindent\textbf{Attention sink in self-attention layer of hybrid MLLM.} 
The ``attention sink''~\cite{xiao2024attentionsink,sun2024massive} is clearly observed in the self-attention layers of Nano, as shown in~\Cref{fig:exp_analyses-attn_mamba_attn_pattern}, where the majority of attention scores are concentrated on the initial few tokens.
This behavior aligns with observations in traditional Transformer models, as exemplified by Qwen's attention maps such as AL24 and AL27.

\noindent\textbf{Decrease of attention to vision tokens across layers.}
Comparing the second row with the first row in~\Cref{fig:exp_analyses-attn_mamba_attn_pattern}, we observe a clear downward trend in attention scores assigned to vision tokens as the layers deepen. This phenomenon is consistent with our broader findings: as the model processes more layers, visual information becomes increasingly redundant and is subsequently deprioritized by instruction and response tokens.

\noindent\textbf{Qualitative results.}
~\Cref{fig:case} illustrates qualitative examples from MCQ, TVG, and VDC tasks. TimeViper accurately answers complex multi-choice video questions, localizes temporal boundaries with an IoU of 0.75, and produces rich, fine-grained video descriptions.

\section{Conclusion}
This work takes an initial step toward understanding and compressing hybrid vision-language models for long videos.
We introduce TimeViper, a Mamba-Transformer hybrid model equipped with TransV, an internal LLM token transfer module that compresses vision information into text tokens.
We reveal that visual information gradually shifts into text tokens as depth of hybrid LLM layer increases, resulting in substantial redundancy among vision tokens in deep layers.
TimeViper can efficiently process hour-long videos while maintaining strong multimodal understanding.
TimeViper achieves promising performance for long video understanding across multi-choice video QA, temporal video grounding, and detailed video captioning.

\clearpage
\appendix
\section*{Appendix}
\addcontentsline{toc}{section}{Supplementary Materials}

\section{Limitations}
First, our current performance still falls short of the SOTA models due to limited training data and insufficient model training. 
Second, while TransV enables processing over 10,000 frames, the model has not been trained on videos of such duration.

\section{Experimental Setups}
\noindent\textbf{Implementation details.}
When training attention-based dropping in multi-turn dialogue scenarios, the attention distribution is computed using the last token of the final instruction as the query.
For temporal video grounding data, we incorporate a time-aware prompt~\cite{li2024videochatflash}: \textit{``The video lasts for \{\} seconds, and \{\} frames are uniformly sampled from it.''} 
During training, we randomly sample one instruction from a pool of 15 manually constructed task prompts, such as: \textit{``From the video, locate the portion that aligns with the textual query {}, and output the start and end timestamps in seconds. The output format of the predicted timestamp should be like: 'start to end' seconds. A specific example is : 12.0 to 20.0 seconds''}. 
We do not use a system prompt, but we retain the BOS token to act as an attention sink~\cite{xiao2024attentionsink}.

\noindent\textbf{Training data summary.}
Our training pipeline adopts a two-stage strategy.
We summarize the training data in~\Cref{tab:data_recipe}.
Specifically, in the first image-text alignment stage, we utilize 3 million images randomly sampled from the CC12M dataset~\cite{changpinyo2021cc12m}, paired with captions sourced from PixelProse~\cite{singla2024pixelprose}.
In the second video instruction tuning stage, we assemble a composite dataset to enhance MLLM's video understanding and timestamp prediction capabilities, comprising: 
(1) 1.3M samples from LLaVA-Video~\cite{zhang2025llavavideo};
(2) 253K data from Kinetics400~\cite{carreira2017k400} and WebVid~\cite{bain2021webvid} that are recaptioned with GPT-4o or Gemini by ShareGemini~\cite{sharegemini} and ShareGPT-4~\cite{chen2024sharegpt4video};
(3) 100K samples from ET-Instruct~\cite{liu2024etbench}; 
(4) 112K samples from VideoGPT-Plus~\cite{Maaz2024VideoGPT+};
(5) 11K samples from LongVid~\cite{li2024videochatflash} and MovieChat~\cite{song2024moviechat};
(6) 26K dense video captioning (DVC) samples aggregated from ActivityNet~\cite{caba2015activitynet}, COIN~\cite{tang2019coin}, HiREST~\cite{zala2023hirest}, ViTT~\cite{huang2020vitt}, and YouCook2~\cite{zhou2018youcook2};
and (7) 250K temporal video grounding (TVG) samples~\cite{wang2025timer1} from YT-Temporal~\cite{yang2023vid2seq}, DiDeMo~\cite{anne2017didemo}, QuerYD~\cite{oncescu2021queryd}, InternVid~\cite{wang2024internvid}, and HowTo100M~\cite{miech2019howto100m}. 
We obtain grounding data with annotations from VTG-IT~\cite{guo2025vtgllm}, TimeIT~\cite{ren2024timechat}, TimePro~\cite{zeng2025timesuite}, HTStep~\cite{afouras2023htstep}, and LongVid~\cite{li2024videochatflash}.
This data collection process yields 339K temporal grounding samples. 
To ensure data quality, we apply a simple cleaning protocol to the TVG data. 
Specifically, we filter out coarse-grained samples where the ground truth duration exceeds 30 seconds or spans more than one-third of the total video length. 
We also discard invalid entries containing out-of-bound timestamps.
Consequently, our TVG training data remains 250K.

\begin{figure}[t]
  \centering  \includegraphics[width=1\linewidth]{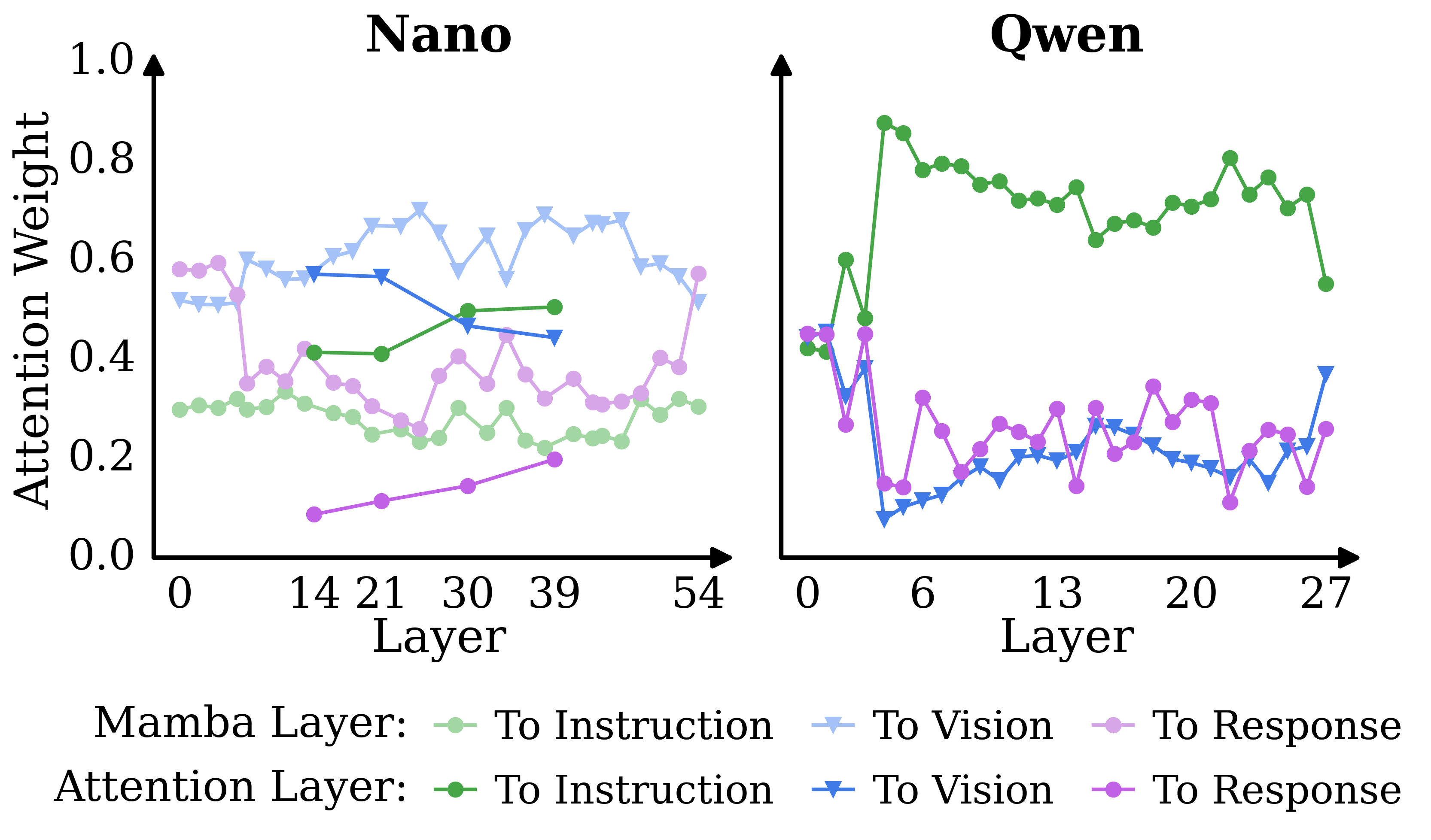}
  \caption{
  Comparison of average attention scores across all layers in Nanov2 and Qwen2.5. 
  The visualization shows both attention and Mamba layers for Nano, and attention layers for Qwen. 
  For Mamba layers, we normalize each row of the attention scores using the $L_1$ norm so that all values fall within the range $[0,1]$.
  }
  \label{fig:exp_analyses-comp_qwen_nano_attn}
\vspace{-8pt}
\end{figure}

\begin{table*}[thbp]
    \caption{{Data recipe.} Overview of the datasets used in our two-stage training pipeline.}
    \label{tab:data_recipe}
    \renewcommand{\arraystretch}{1.3} 
    \centering
    \small
    \renewcommand{\tabularxcolumn}[1]{m{#1}} 
    \begin{tabularx}{\textwidth}{l X} 
        \toprule
        \rowcolor{lightgray}
        \multicolumn{2}{l}{\textit{\textbf{Stage 1: Projector Alignment}}} \\
        \hline
        Image caption data (3M) & CC12M (3M)~\cite{changpinyo2021cc12m} with PixelProse captions~\cite{singla2024pixelprose} \\
        \hline \hline
        \rowcolor{lightgray}
        \multicolumn{2}{l}{\textit{\textbf{Stage 2: Video Instruction-Tuning}}} \\
        \hline
        Image instruction data (2.8M) & LLaVA-OneVision (2.8M)~\cite{lillavaov}; \\
        \hline
        Video instruction data (1.8M) & LLaVA-Video (1.3M)~\cite{zhang2025llavavideo}; {Kinetics400 \& WebVid (253K)}~\cite{carreira2017k400,bain2021webvid} (recaptioned via ShareGemini~\cite{sharegemini} \& ShareGPT-4~\cite{chen2024sharegpt4video}); {VideoGPT-Plus (112K)}~\cite{Maaz2024VideoGPT+}; {ET-Instruct (100K)}~\cite{liu2024etbench}; {LongVid~\cite{li2024videochatflash} \& MovieChat (11K)}~\cite{song2024moviechat} \\
        \hline
        Dense video captioning (26K) & ActivityNet~\cite{caba2015activitynet}, COIN~\cite{tang2019coin}, HiREST~\cite{zala2023hirest}, ViTT~\cite{huang2020vitt}, YouCook2~\cite{zhou2018youcook2} \\
        \hline
        Temporal video grounding (250K) & YT-Temporal~\cite{yang2023vid2seq}, DiDeMo~\cite{anne2017didemo}, QuerYD~\cite{oncescu2021queryd}, InternVid~\cite{wang2024internvid}, HowTo100M~\cite{miech2019howto100m} (Annotated by VTG-IT~\cite{guo2025vtgllm}, TimeIT~\cite{ren2024timechat}, TimePro~\cite{zeng2025timesuite}, HTStep~\cite{afouras2023htstep}, LongVid~\cite{li2024videochatflash}) \\
        \bottomrule
    \end{tabularx}
\end{table*}

\begin{table*}[t]
\centering
\caption{Performance of applying TransV to Qwen2.5 and Nano.
}
\label{tab:supple_qwen_transv}
\resizebox{\textwidth}{!}{%
\begin{tabular}{l c c c c c cc c c c}
\toprule
\multirow{2}{*}{\textbf{Model}} & 
\multirow{2}{*}{\textbf{LLM}} & 
\multirow{2}{*}{\textbf{$>$10K frame input}} & 
\textbf{MVBench} & 
\textbf{LongVideoBench} & 
\textbf{MLVU} & 
\multicolumn{2}{c}{\textbf{VideoMME}} & 
\textbf{LVBench} & 
\textbf{Charades-STA} & 
\textbf{VDC} \\
\cmidrule(lr){7-8}
& &  & avg.acc & val & M-Avg & overall & long & avg.acc & mIoU & avg.acc \\
\midrule
\textbf{Qwen2.5-7B (ours)} & Qwen2.5-7B~\cite{qwen2.5} & \ding{55}  & 57.6 & 55.4 & 64.9 & 56.6 & 48.7 & 36.6 & 40.8 & 42.0 \\ 
\textbf{Qwen2.5-7B (ours w/ TransV)} & Qwen2.5-7B~\cite{qwen2.5} & \textcolor{blue}{\ding{51}}  & 55.7 & 53.7 & 63.3 & 55.7 & 47.4 & 37.0 & 38.7 & 40.7 \\ 
\midrule
\textbf{TimeViper (ours)} & Nanov2-9B~\cite{basant2025nanov2} & \ding{55} & 57.2 & 54.1 & 65.6 & 58.8 & 48.8 & 35.5 & 40.5 & 39.7 \\ 
\textbf{TimeViper (ours w/ TransV)} & Nanov2-9B~\cite{basant2025nanov2} & \textcolor{blue}{\ding{51}} & 56.2 & 52.0 & 63.1 & 56.9 & 48.2 & 35.6 & 37.9 & 39.1 \\ 
\bottomrule
\end{tabular}%
}

\end{table*}

\section{Main Results}
\noindent\textbf{TransV can be also applied to Qwen2.5.}
Qwen can also use TransV to handle ultra-long sequences, as shown in~\Cref{tab:supple_qwen_transv}. 
We observe that on LVBench, this even brings a +0.4 improvement. 
However, the performance drop on VDC is more severe for Qwen than for Nano. 
For example, Qwen drops from 42.0 to 40.7 (a decrease of 1.3 points), whereas Nano drops from 39.7 to 39.1 (a decrease of 0.6 points).

\section{Qualitative Results}

We define average attention scores used in both Mamba-2 and self-attention layers to analyze attentions received by different types of tokens.

\noindent\textbf{Average attention score computation.} We adopt the category-level attention score definition from LLaVA-Mini~\cite{zhang2025llavamini}.
Tokens are grouped into instruction, vision, and response categories:
$\mathcal{T}_\text{ins}$, $\mathcal{T}_\text{vis}$, and $\mathcal{T}_\text{res}$.
Let $a_{ij}$ denote the attention score from token $t_i$ to token $t_j$, averaged over all attention heads.
For two token categories $\mathcal{A}, \mathcal{B} \in \{\mathcal{T}_\text{ins}, \mathcal{T}_\text{vis}, \mathcal{T}_\text{res}\}$,
we define their category-level attention score as:
\begin{equation}
\text{Attn}(\mathcal{A} \!\to\! \mathcal{B})
=
\frac{
\sum_{t_i \in \mathcal{A}} \sum_{t_j \in \mathcal{B}} a_{ij}
}{
\left|\left\{\, t_i \in \mathcal{A} \;\big|\; \sum_{t_j \in \mathcal{B}} a_{ij} > 0 \right\}\right|
}.
\label{eq:avg_attn}
\end{equation}

\noindent The denominator counts the number of tokens in $\mathcal{A}$ that attend to any token in $\mathcal{B}$ with non-zero weight, ensuring that tokens masked by the causal attention mask are excluded.

\vspace{4pt} \noindent In~\Cref{fig:exp_analyses-comp_qwen_nano_attn}, we analyze the overall attention scores from the entire sequence
$\mathcal{T} = \mathcal{T}_\text{ins} \cup \mathcal{T}_\text{vis} \cup \mathcal{T}_\text{res}$
to a target category $\mathcal{B}$.  To ensure equal contribution from each category, we compute the arithmetic mean of the scores:
\begin{equation}
\begin{aligned}
\text{Attn}(\mathcal{T} \!\to\! \mathcal{B})
=
\frac{1}{3}
\big(
& \text{Attn}(\mathcal{T}_\text{ins} \!\to\! \mathcal{B}) \\
+ & \text{Attn}(\mathcal{T}_\text{vis} \!\to\! \mathcal{B}) \\
+ & \text{Attn}(\mathcal{T}_\text{res} \!\to\! \mathcal{B})
\big).
\end{aligned}
\end{equation}

\noindent\textbf{Hybrid MLLMs preserve stronger attention to vision tokens.} 
To quantify model behavior, we compute the average attention received by instruction, vision, and response tokens across all layers.
As shown in~\Cref{fig:exp_analyses-comp_qwen_nano_attn}, Qwen rapidly down-weights vision tokens after the early layers, instead favoring instruction and response tokens.
In contrast, Nano maintains noticeably higher attention to vision tokens throughout the network.
These findings suggest that the hybrid model is more effective at attending to visual information than the Transformer-based architecture.

\begin{figure*}[t]
  \centering

  \begin{subfigure}{\linewidth}
    \centering
    \includegraphics[width=0.83\linewidth,keepaspectratio]{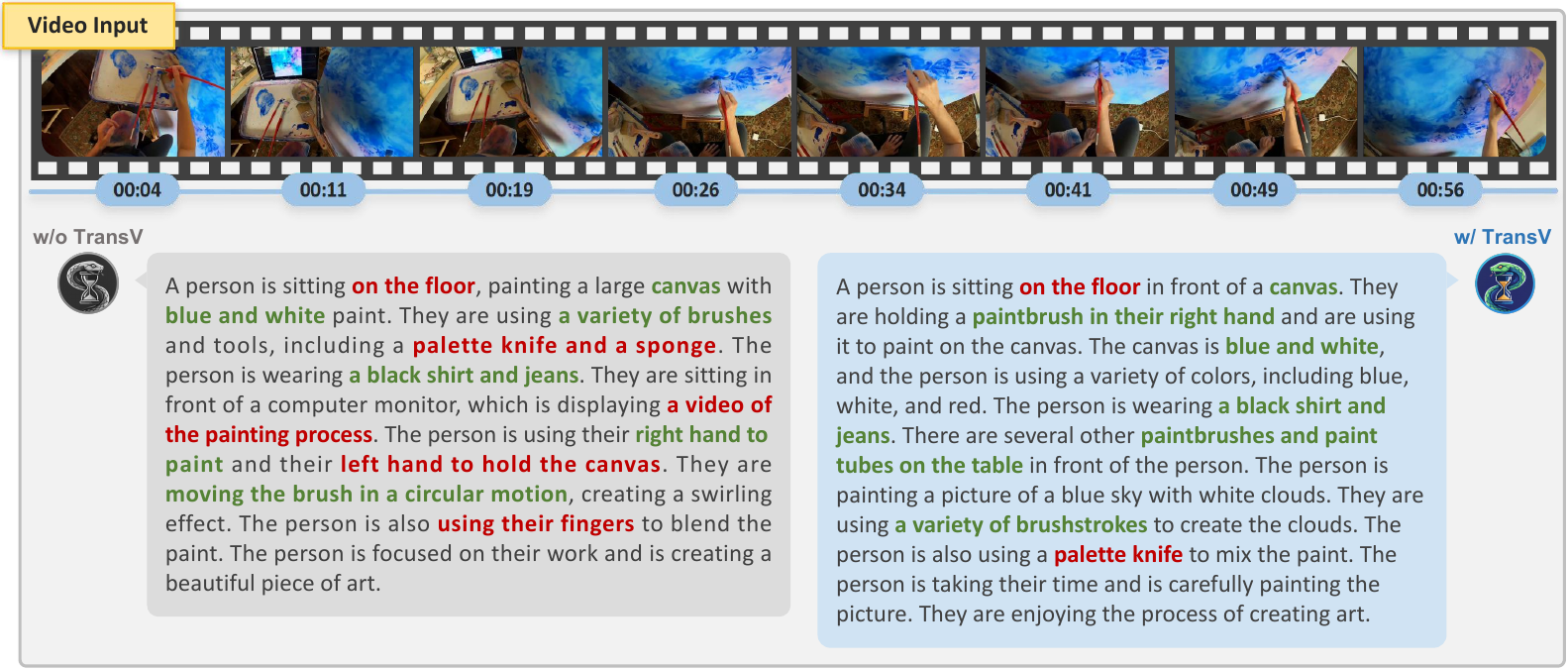}
    \caption{Qualitative results on VDC.}
    \label{fig:case_vdc}
  \end{subfigure}

  \vspace{6pt} 

  \begin{subfigure}{\linewidth}
    \centering
    \includegraphics[width=0.83\linewidth,keepaspectratio]{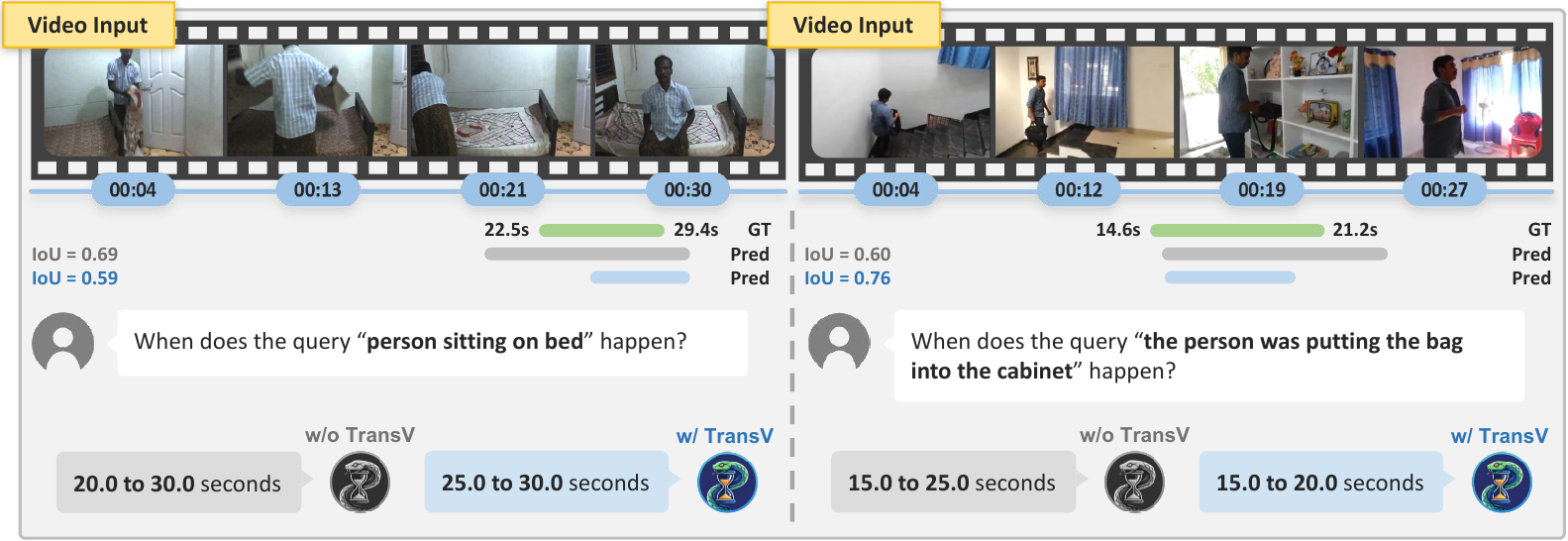}
    \caption{Qualitative results on Charades.}
    \label{fig:case_charades}
  \end{subfigure}

  \vspace{6pt}

  \begin{subfigure}{\linewidth}
    \centering
    \includegraphics[width=0.83\linewidth,keepaspectratio]{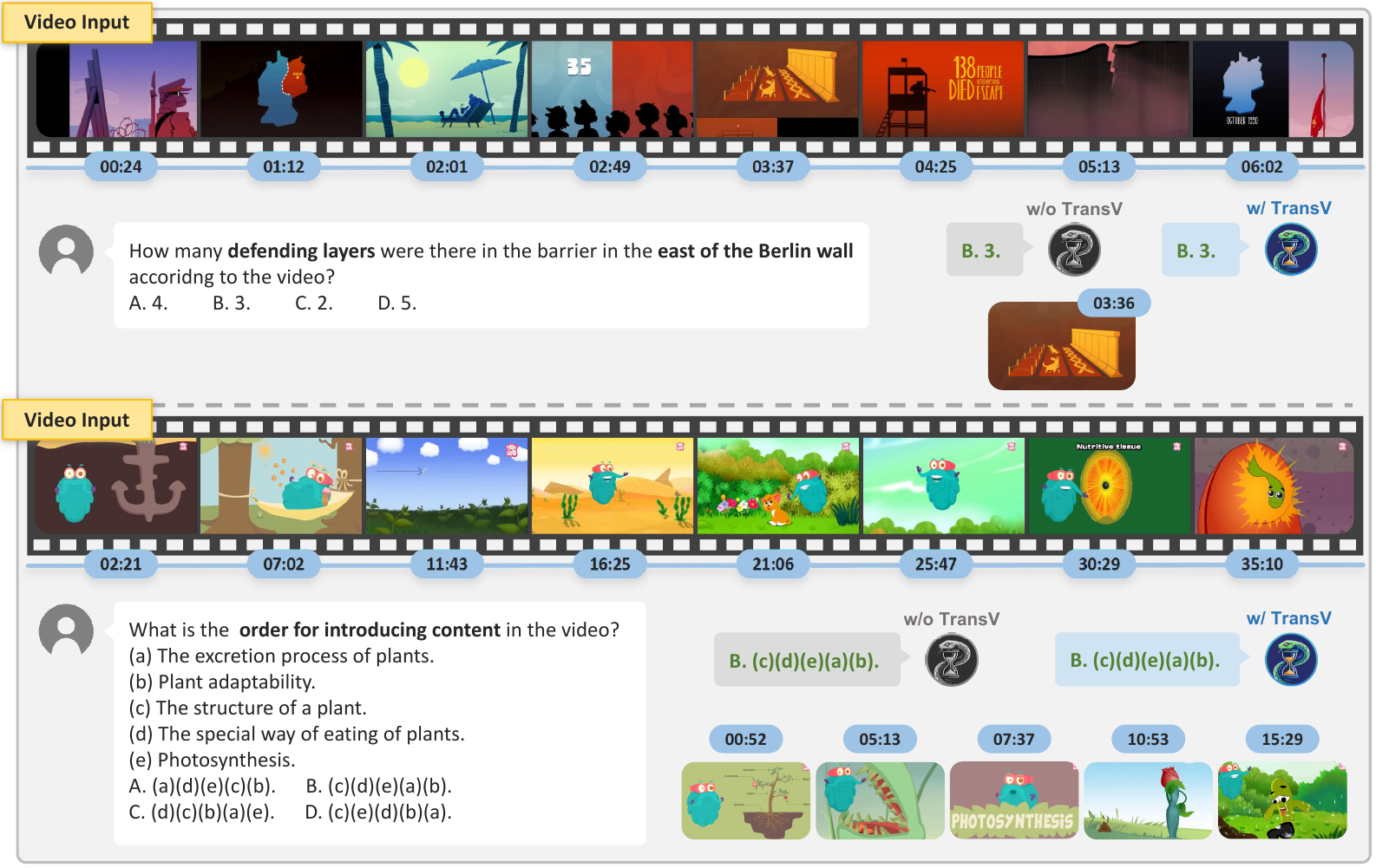}
    \caption{Qualitative results on VideoMME.}
    \label{fig:case_videomme}
  \end{subfigure}

  \caption{Qualitative results on three benchmarks.}
  \label{fig:three_vertical_cases}
\end{figure*}

\noindent\textbf{Qualitative results on VideoMME.}
~\Cref{fig:case_videomme} provides qualitative examples illustrating the effectiveness of TransV on the MCQ task.
In the first case (top row), the query requires retrieving fine-grained visual information, specifically, the defending layers about the Berlin Wall. 
The compressed model, TimeViper w/ TransV, successfully attends to the critical frame at 03:36, which clearly depicts the structure of the wall, enabling it to select the correct answer. 
This example highlights TransV's capability to identify and attend to key visual cues within long videos.
In the second case (bottom row), the query requires long-term temporal reasoning to infer the chronological order of topics in a biology lecture. 
The correct answer relies on aggregating information across the entire video duration. 
In this case, a model must accurately align the textual concepts (e.g., structure, photosynthesis) with their corresponding temporal segments (00:52, 07:37, etc.), 
As shown in the visual evidence, TimeViper w/ TransV correctly deduces the sequence (c)-(d)-(e)-(a)-(b), demonstrating its capability in modeling global temporal dependencies and understanding narrative structure.

\noindent\textbf{Qualitative results on Charades.}
For temporal video grounding results shown in~\Cref{fig:case_charades}, we observe that incorporating the compression module yields only minimal changes. Both the original and compressed models correctly interpret timestamps and successfully localize video segments that correspond to the natural language query.

\noindent\textbf{Qualitative results on VDC.}
~\Cref{fig:case_vdc} presents the qualitative comparison for the video detailed captioning task. 
In the generated captions, green text denotes accurate visual details, while red text indicates hallucinations or factual errors.
In the first example showing a person painting (top row), the baseline model suffers from severe object hallucination, fabricating elements such as a ``sponge'' which are absent in the video. 
Surprisingly, TimeViper w/ TransV generates more faithful description, accurately recognizing specific objects like ``paintbrushes''. 
This suggests that compression may help reduce hallucination by filtering out irrelevant or misleading visual information.
In the second example (bottom row), the compression module TransV largely retains the original model behavior where there are both correct and incorrect captions.



\clearpage
{
\bibliographystyle{ieeenat_fullname}
\bibliography{main}
}

\end{document}